\documentclass[sigconf, nonacm]{acmart}

\newcommand\vldbdoi{XX.XX/XXX.XX}

\usepackage{float}
\usepackage{footnote}
\usepackage{graphicx} 
\usepackage{placeins}
\usepackage{enumitem}
\usepackage{subcaption}
\usepackage{multirow}
\usepackage{color, soul}

\usepackage{wrapfig}
\begin{document}
\title{KeyVideoLLM:  Towards Large-scale Video Keyframe Selection}
\author{Hao Liang$^{\dagger\spadesuit}$, Jiapeng Li$^{\dagger\heartsuit}$, Tianyi Bai$^\diamondsuit$, Xijie Huang$^\spadesuit$, Linzhuang Sun$^\S$, Zhengren Wang$^\spadesuit$, Conghui He$^\diamondsuit$, Bin Cui$^\spadesuit$, Chong Chen$^{\clubsuit*}$, Wentao Zhang$^{\spadesuit*}$}
\affiliation{
~~~~~$^\spadesuit$Peking University~~~~~$^\heartsuit$The Open University of China~~~~~$^\clubsuit$Huawei Cloud BU\\
~~~~~$^\diamondsuit$Shanghai AI Laboratory~~~~~$^\S$University of Chinese Academy of Sciences
 }
\affiliation{
hao.liang@stu.pku.edu.cn, jasper\_li@alumni.pku.edu.cn, chenchong55@huawei.com wentao.zhang@pku.edu.cn,
}





\begin{abstract}
Recently, with the rise of web videos, managing and understanding large-scale video datasets has become increasingly important. Video Large Language Models (VideoLLMs) have emerged in recent years due to their strong video understanding capabilities. However, training and inference processes for VideoLLMs demand vast amounts of data, presenting significant challenges to data management, particularly regarding efficiency, robustness, and effectiveness. In this work, we present KeyVideoLLM, a text-video frame similarity-based keyframe selection method designed to manage VideoLLM data efficiently, robustly, and effectively. Specifically, KeyVideoLLM achieves a remarkable data compression rate of up to \textbf{60.9 times}, substantially lowering disk space requirements, which proves its high efficiency. 
Additionally, it maintains a \textbf{100\%} selection success rate across all video formats and scales, enhances processing speed by up to \textbf{200} times compared to existing keyframe selection methods, and does not require hyperparameter tuning. 
Beyond its outstanding efficiency and robustness, KeyVideoLLM further improves model performance in video question-answering tasks during both training and inference stages. Notably, it consistently achieved the state-of-the-art (SoTA) experimental results on diverse datasets.
\end{abstract}
\maketitle
\begingroup
\renewcommand\thefootnote{}\footnote{\noindent
$^\dagger$ The first two authors have equal contributions. \\
$*$ Corresponding Authors.
}
\addtocounter{footnote}{-1}
\endgroup



\begin{figure}[htbp]
    \centering
    \begin{subfigure}[b]{0.47\textwidth}
        \centering
        \includegraphics[width=\textwidth]{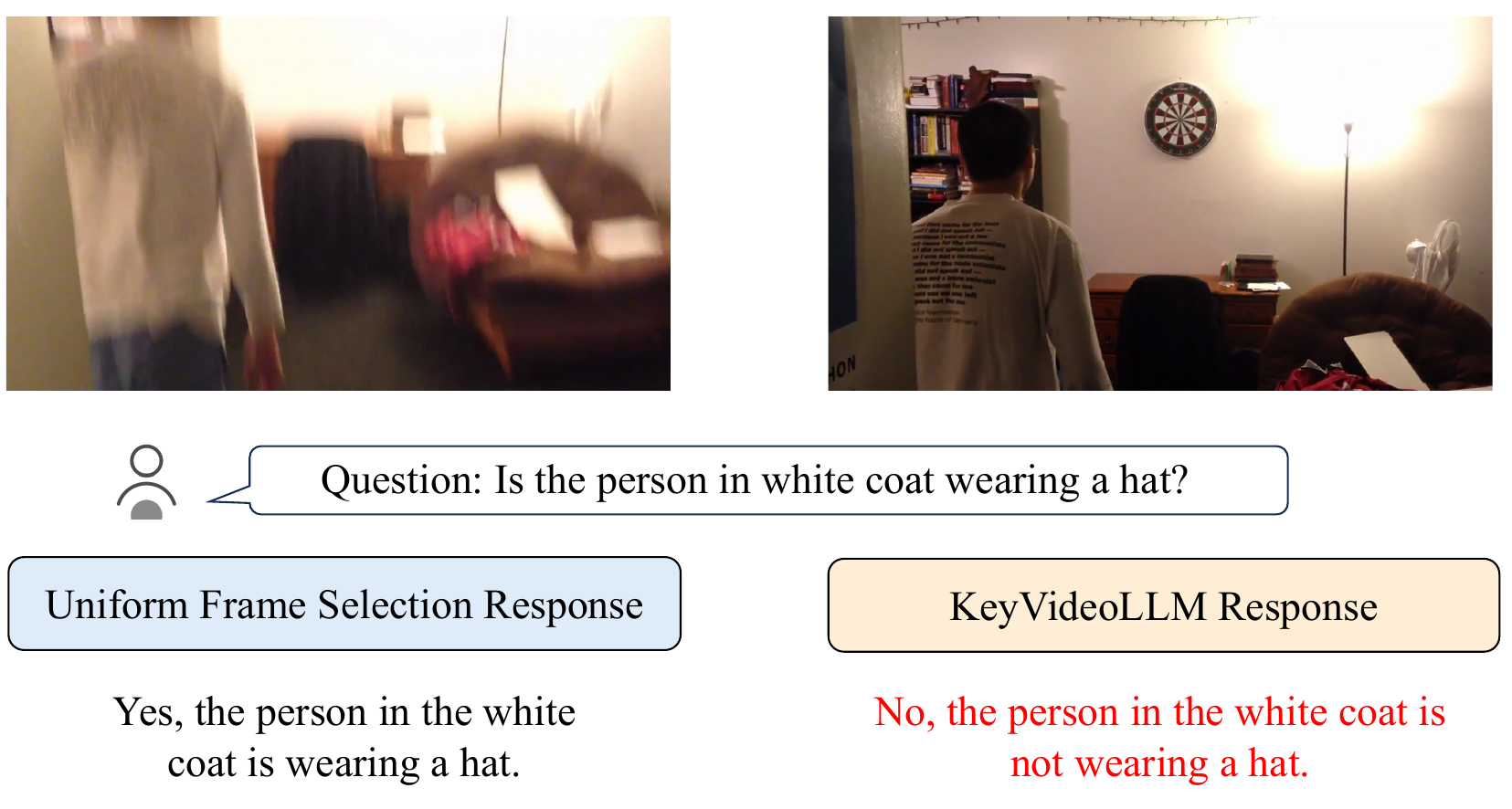}
        \caption{Selected frames from Uniform frame selection, KeyVideoLLM, and corresponding answers.}
        \label{fig:Face_1}
    \end{subfigure}
    \vfill
    \begin{subfigure}[b]{0.47\textwidth}
        \centering
        \includegraphics[width=\textwidth]{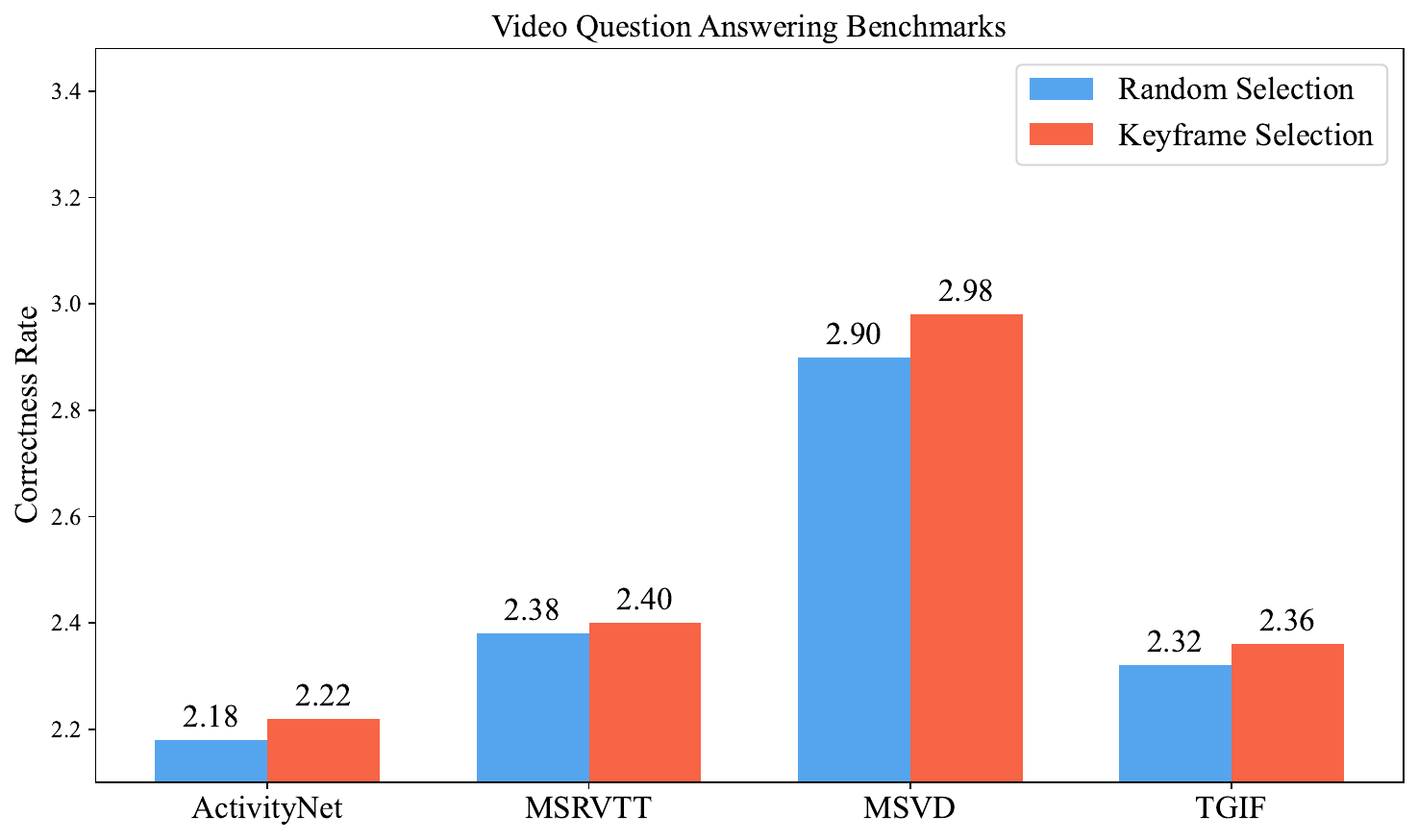}
        \caption{Performance of Uniform and KeyVideoLLM on Video Question-Answering Benchmarks.}
        \label{fig:Face_2}
    \end{subfigure}
    \caption{In (a), uniform frame selection often results in images that lack the information required to answer the question. In contrast, KeyframeLLM ensures the frames contain the necessary information. In (b), KeyVideoLLM improves the performance of VideoLLM across all benchmarks.}
    \label{fig:Face}
\end{figure}
\section{Introduction}

In recent years, with the rapid advancements in large language models (LLMs)~\cite{chatgpt, llama} and multimodal large language models (MLLMs)~\cite{zhao2023survey,wu2023multimodal}, data management has become a crucial aspect of these technologies~\cite{fernandez2023large, trummer2023bert, chen2023lingua, miao2024demystifying, nie2023flexmoe}. At the same time, \citet{bai2024survey} also demonstrates that data processing, selection, and management can significantly influence the performance of MLLMs.

Among MLLMs, VideoLLMs achieve competitive performance in traditional multimodal tasks such as visual recognition~\cite{zhang2024vision}, video understanding~\cite{xu2021vlm, tang2023video}, and action recognition~\cite{internvideo2}. Moreover, their excellent language understanding capabilities enable strong performance in text-rich tasks, such as video question-answering~\cite{hu2024bliva} and video-centric dialogues~\cite{internvideo2}.

Most existing VideoLLMs focus on modifying model architecture to utilize information from multiple modalities~\cite{internvideo, internvideo2, video-chatgpt, video-llama, videochat}. While model effectiveness is crucial, data also significantly impacts the success of VideoLLMs. For instance, \citet{videochat2, internvideo2} demonstrate that higher-quality training data can enhance the performance of VideoLLMs. Additionally, \citet{fernandez2023large} indicates that LLMs can disrupt data management due to their massive data requirements. However, current video data selection methods primarily emphasize video quality, captions, and video-caption alignment, often resulting in redundant datasets. These methods neglect the importance of efficient and robust data management and face the following three key challenges:

\textbf{C1. Low Efficiency.} Due to the large storage requirements of video data, massive training datasets often occupy substantial storage space, ranging from several hundred gigabytes to tens of terabytes~\cite{internvideo, internvideo2, video-llava}. 
Additionally, the common practice of using random or uniform frame selection during training leads to considerable data waste. 
This inefficiency not only increases storage needs but also hinders the model's ability to learn from the most relevant and informative content within the videos.

\textbf{C2. Low Robustness.} Existing keyframe selection methods are sensitive to hyperparameters. For instance, Katna~\cite{katna} and DSNet~\cite{DSNet} are two previous SoTA methods that require extensive hyperparameter tuning. Moreover, the experimental results in Table~\ref{tab:Success_Rate} demonstrate their very low success rates on short videos. Additionally, Table~\ref{tab:Extract_Speed} reveals that their keyframe selection speeds are relatively slow.

\textbf{C3. Poor Effectiveness.} Typically, VideoLLMs employ uniform or random frame selection methods during the training stage and uniform frame selection methods during the inference stage~\cite{video-llava, internvideo, internvideo2}. 
These uniform or random selection methods do not consider the relevance of frames to the questions and answers. 
As illustrated on the left of Figure \ref{fig:Face_1}, the uniform selection method fails to select frames relevant to the question, resulting in incorrect answers.

To address these issues, we propose KeyVideoLLM. KeyVideoLLM leverages the power of deep learning models to perform precise keyframe selection, ensuring that the selected frames are highly relevant to the given query and response based on text-video frames similarity scores. 
Specifically, KeyVideoLLM performs precise keyframe selection which is extremely efficient in both data usage and disk storage. 
Additionally, it leverages the strong parallel computing capabilities of GPUs and employs a coarse-to-fine keyframe selection process, resulting in very fast selection speeds and high success rates with almost no hyperparameters required. 
We then use KeyVideoLLM for VideoLLMs training and inference to improve the model's effectiveness.
In the training phase, we use KeyVideoLLM based on answer and question-answer similarities to select keyframes more relevant to the answer or the question-answer pair. 
As shown in Figure \ref{fig:Face_1}, selecting more relevant frames helps improve model performance, resulting in correct answers.
In the inference phase, 
we employ KeyVideoLLM based on the question to select frames related to the question. As shown in Figure \ref{fig:Face_2}, more relevant keyframes result in more effective VideoLLMs. 

The core contributions of this paper are summarized as follows:





\begin{figure*}[ht]
  \includegraphics[width=16cm]{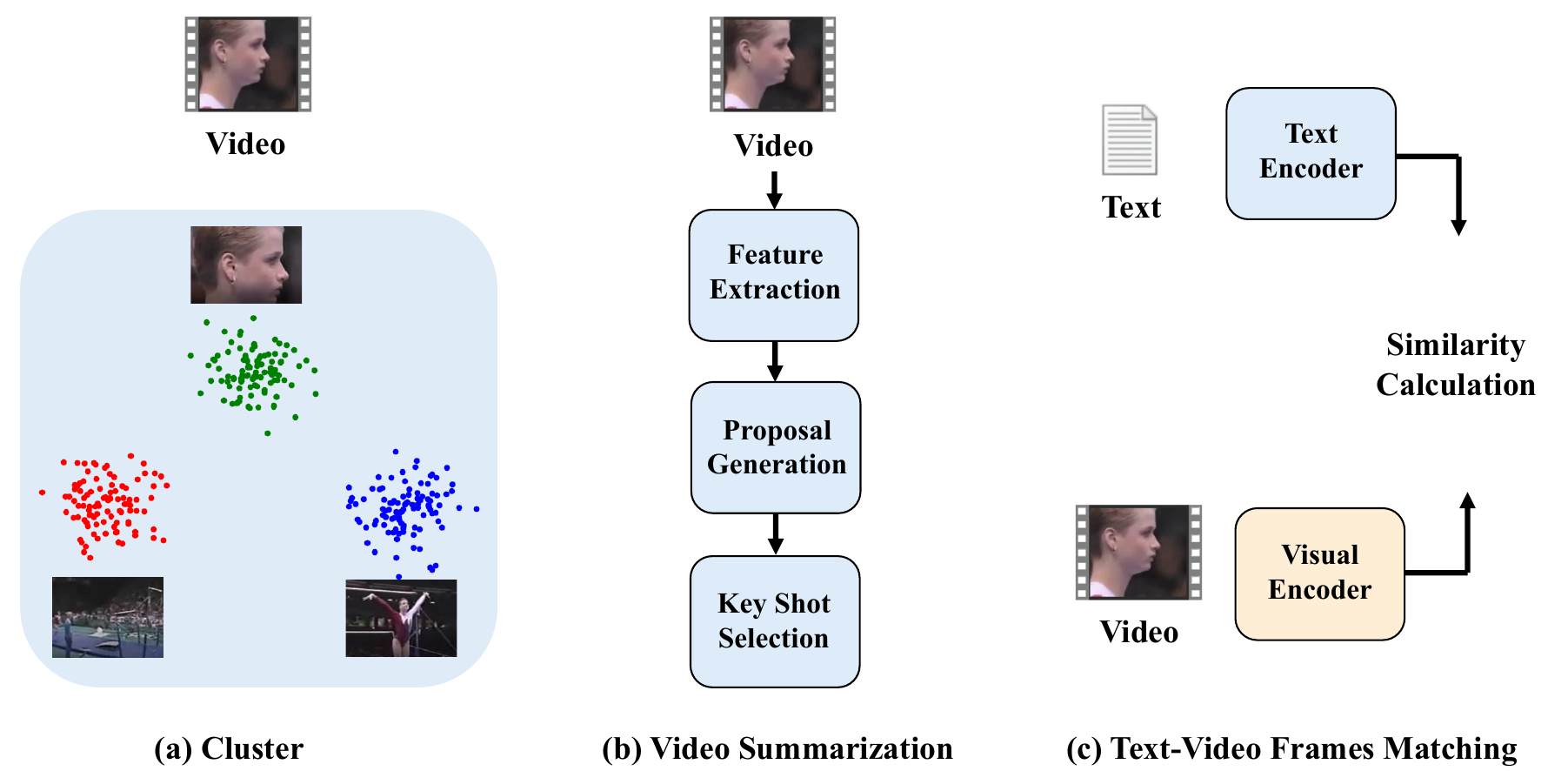}
  \caption{Comparison of three methods for keyframe selection. To the best of our knowledge, this is the first study to select video frames using text-video frames matching for VideoLLMs.}
  \label{fig:method_compare}
\end{figure*}

\begin{itemize}
    \item \textbf{New Perspective.} Low efficiency and low robustness are significant impediments to the practical adoption of keyframe selection methods. To the best of our knowledge, this study represents the first attempt to address these challenges from a data management perspective.
    \item \textbf{New Method.} We propose KeyVideoLLM, the first text-video frame similarity-based keyframe selection method. Based on the proposed text-video frames similarity scores, KeyVideoLLM can manage VideoLLM data efficiently, robustly, and effectively.
    \item \textbf{SoTA Performance.} 
    \textbf{(1)\textit{High Efficiency.}} KeyVideoLLM is highly efficient, achieving a data compression rate of up to 60.9 times, significantly reducing disk usage. As shown on the right side of Figure \ref{fig:Face_1}, it effectively selects frames relevant to the question, mitigating the waste of video data.
    \textbf{(2)\textit{High Robustness.}} KeyVideoLLM can achieve selection speeds up to 200 times faster per video. It also achieves the highest keyframe selection success rate compared to previous keyframe selection methods. Unlike existing methods, KeyVideoLLM does not require additional hyperparameter tuning, demonstrating its robustness.
    \textbf{(3)\textit{Effectiveness in Training and Inference Stage.}} Our answer and question-answer-based KeyframeLLM improve the performance of VideoLLMs during the training stage compared to uniform frame selection, such as Katna~\cite{katna} and DSNet~\cite{DSNet}. Besides, our question-based selection method further enhances the performance of VideoLLMs during the inference stage compared to uniform selection, achieving SoTA performance.
\end{itemize}


\section{Related Work}
\paragraph{\textbf{Video Multimodal Models.}} Recently, inspired by the remarkable understanding capabilities of LLMs and pre-trained models, researchers have started using LLMs to understand videos, achieving SoTA results~\cite{videochat2,video-llama,minigpt4,internvideo2,video-llava}. VideoLLaMA~\cite{video-llama} is one of the pioneering studies in VideoLLMs, utilizing a visual encoder and a video Q-Former projector to understand videos. However, due to its Q-Former structure, the computational cost is very high. To address this, subsequent works~\cite{video-llava} adopted the LLaVA~\cite{llava, llava1.5} MLP structure, significantly reducing computational costs while still achieving SoTA performance. Similarly, MiniGPT4Video~\cite{minigpt4} uses an MLP adapter for efficient training.

Another notable series of models includes VideoChat, VideoChat2, InternVideo, and InternVideo2~\cite{videochat, videochat2, internvideo, internvideo2}. These models utilize an enormous amount of data to train a transformer-structured adapter, achieving SoTA performance. By leveraging large-scale datasets and advanced transformer architectures, these models excel in comprehending and processing multimodal video content, further pushing the boundaries of video understanding capabilities.

\paragraph{\textbf{Keyframe Selection for Video Multimodal Models.}}
VideoLLMs often integrate frame encoding techniques to mitigate resource overhead and streamline training durations. Most VideoLLMs~\cite{video-llava,internvideo, pllava,video-llama} employ a uniform sampling methodology to select a fixed number of frames. This approach is also used during the testing phase of InternVideo2~\cite{internvideo2} and VideoChat2~\cite{videochat2}. However, during the training phase, these models opt for random frame selection within each time interval. Some models~\cite{semantically} leverage pre-existing compressed video methodologies, such as those facilitated by ffmpeg, to select frames for training purposes. Katna~\cite{katna}, a frame selection method incorporating machine learning techniques, is employed by VideoChatGPT for frame selection. Additionally, certain architectures~\cite{minigpt4,llmsmeet,koala} incorporate supplementary modules aimed at reducing the token count encoded per keyframe, thereby improving computational efficiency and avoiding input token constraints. Video-LaVIT~\cite{video-lavit} employs a fusion of keyframes and motion vectors to tokenize video data. These diverse strategies for keyframe management not only impact the computational dynamics of model training and inference but also significantly influence the resultant quality metrics of video-centric LLMs.
\paragraph{\textbf{Data-Centric LLMs and Data Selection Methods}}
The advent of LLMs has led to a substantial increase in the volume of training data~\cite{llama, openai2023gpt}. VideoLLMs face even higher storage and computational costs due to the vast amount of data and substantial storage space required for video content~\cite{internvideo2}. This increase in data volume also brings new challenges in data management and selection~\cite{bai2024survey}.

LLM-based methods are commonly used in data selection~\cite{bai2024survey}. For instance, \citet{du2023mods} leverage DeBERTa~\cite{he2020deberta} for scoring, retaining high-quality data, and combining it with the k-center greedy algorithm to select diverse data. \citet{chen2023alpagasus} score the accuracy of data using ChatGPT to identify high-quality data. \citet{xu2023rethinking} use GPT-4 to rewrite data to increase its complexity and then streamline it by reducing its variety and improving its quality. \citet{liu2023makes} train two models using ChatGPT-labeled data to score the quality and complexity of the data. \citet{lu2023instag} rely on ChatGPT to tag each instance, defining its complexity and diversity based on these tags. \citet{parkar2024selectllm} first cluster the data, and then use GPT-4 to select high-quality data for each cluster.

\section{Method}
\subsection{Keyframe Selection}
To the best of our knowledge, this study is the first to select video frames using text-video frames matching for training VideoLLMs. We categorize frame selection methods into three distinct categories, as illustrated in Fig.\ref{fig:method_compare}. Here, we first summarize Cluster and Video Summarization-based methods.
\subsubsection{Cluster}
These methods select the best images from each cluster by first preparing the clusters based on histograms. Katna~\cite{katna} is one of the representatives of these methods. It calculates the histograms for each image and adds them to the histogram list. Then, Katna uses K-means clustering on the histograms to identify the label for each image in the cluster and tag images. The K-means method assigns each frame to the cluster where the nearest center point is located, then updates the center point by recalculating the center point of each cluster. Katna repeats these steps until the cluster center converges or reaches the maximum number of iterations. Afterward, Katna selects the best images from every cluster by choosing the image with the lowest blur (high Laplacian) score. However, the effectiveness of such algorithms largely depends on feature selection and parameter settings. Different settings of the hyperparameters have a relatively large impact on the effectiveness of frame selection.
\begin{figure*}[t]
  \includegraphics[width=16cm]{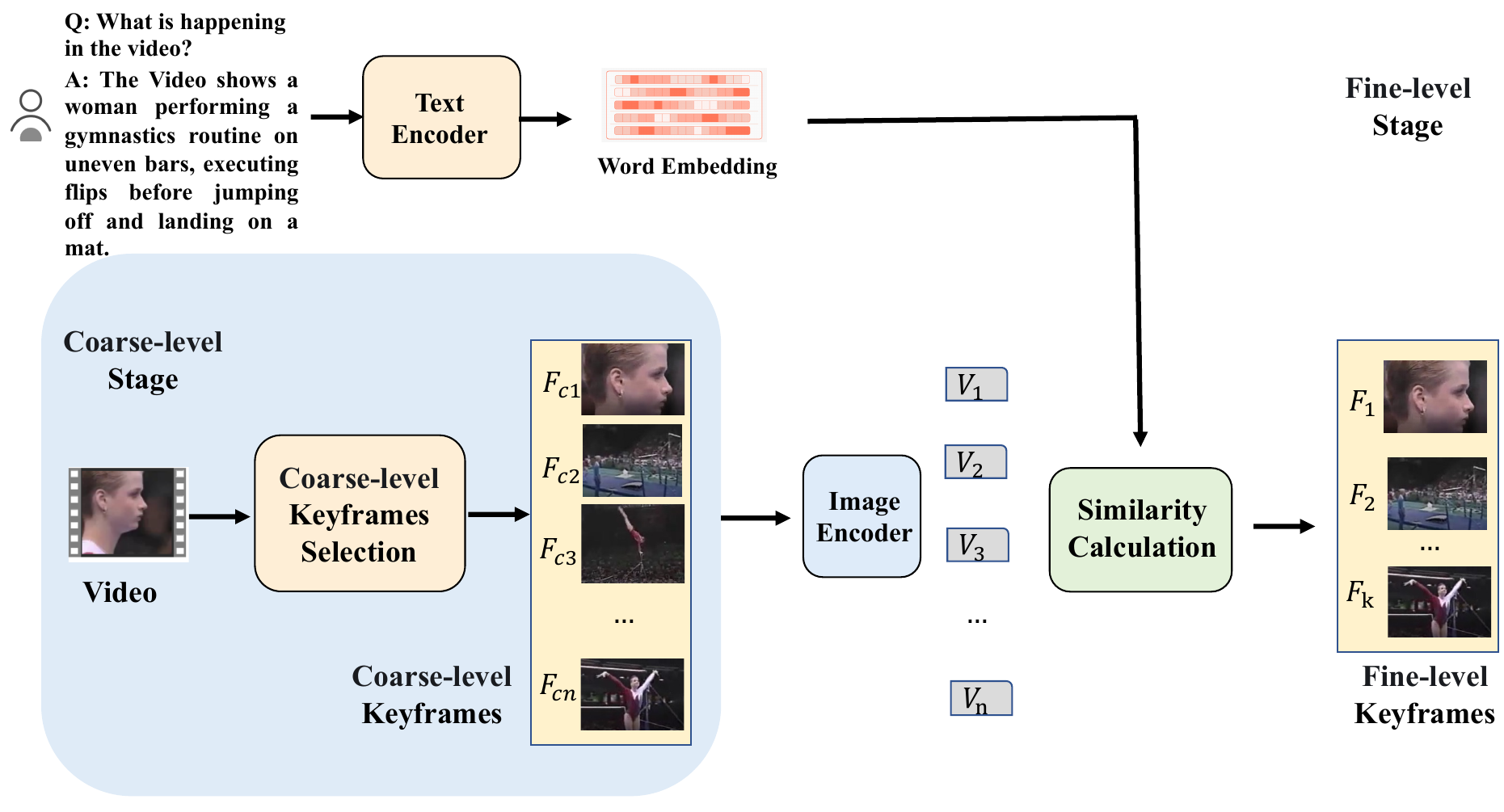}
  \caption{We propose a frame selection method based on text-video frame matching. The method follows a coarse-to-fine framework. We use pre-trained models to select information from text and video frames.}
  \label{fig:clip_coarse_fine}
\end{figure*}

\subsubsection{Video Summarization}
Video summarization technologies aim to create a concise and complete synopsis by selecting the most informative parts of the video content~\cite{Video_summarization_survey_apostolidis2021video}. Existing video summarization methods suffer from dynamic visual context and overfitting problems, which can easily lead to incorrect and incomplete video summaries. DSNet~\cite{DSNet} is one of the representatives of these methods. It consists of feature selection, interest proposal generation, and key shot selection steps. For the feature selection stage, the model selects frame-level features and applies a temporal modeling layer to capture long-range representations. Then, DSNet applies a shared classification and regression module to predict the importance score, center-ness score, and segment boundaries at each temporal location. For testing, segments are refined using the predicted locations and further filtered with non-maximum suppression. Finally, the video summary is generated using a dynamic programming algorithm.

\subsection{Text-Video Frame Similarity Based Keyframe Selection}\label{sec:CLIP-based Keyframe Extraction}
We propose a frame selection method based on text-video frames matching. The method follows a coarse-to-fine framework, as shown in Fig.~\ref{fig:clip_coarse_fine}. 

Given a (video, text) pair, we aim to select frames related to the text content from the video as keyframes. To match semantically similar texts and images, we require a multi-modal embedding space that maximizes the cosine similarity between the keyframe and text embeddings. Inspired by \cite{X-pool}, we use a pre-trained CLIP~\cite{clip} model as a backbone. CLIP~\cite{clip} is a model developed by OpenAI that aligns images with textual descriptions in a shared embedding space.  We use CLIP as the image and text encoder due to its robust ability to learn and represent both visual and textual data in a shared embedding space.

The text encoder must first select appropriate text information to describe a video. In the training stage, we compared two methods for selecting text information. The first method uses only the answers in the conversation, denoted as CLIP-A. The second method uses both the questions and corresponding answers, denoted as CLIP-QA. In the inference stage, only questions are provided, and we use them to select relevant frames, denoted as CLIP-Q. For CLIP-A, CLIP-QA, and CLIP-Q, we use the pre-trained CLIP model's text encoder to map the caption into an embedding space.

\begin{figure*}[t]
  \includegraphics[width=16cm]{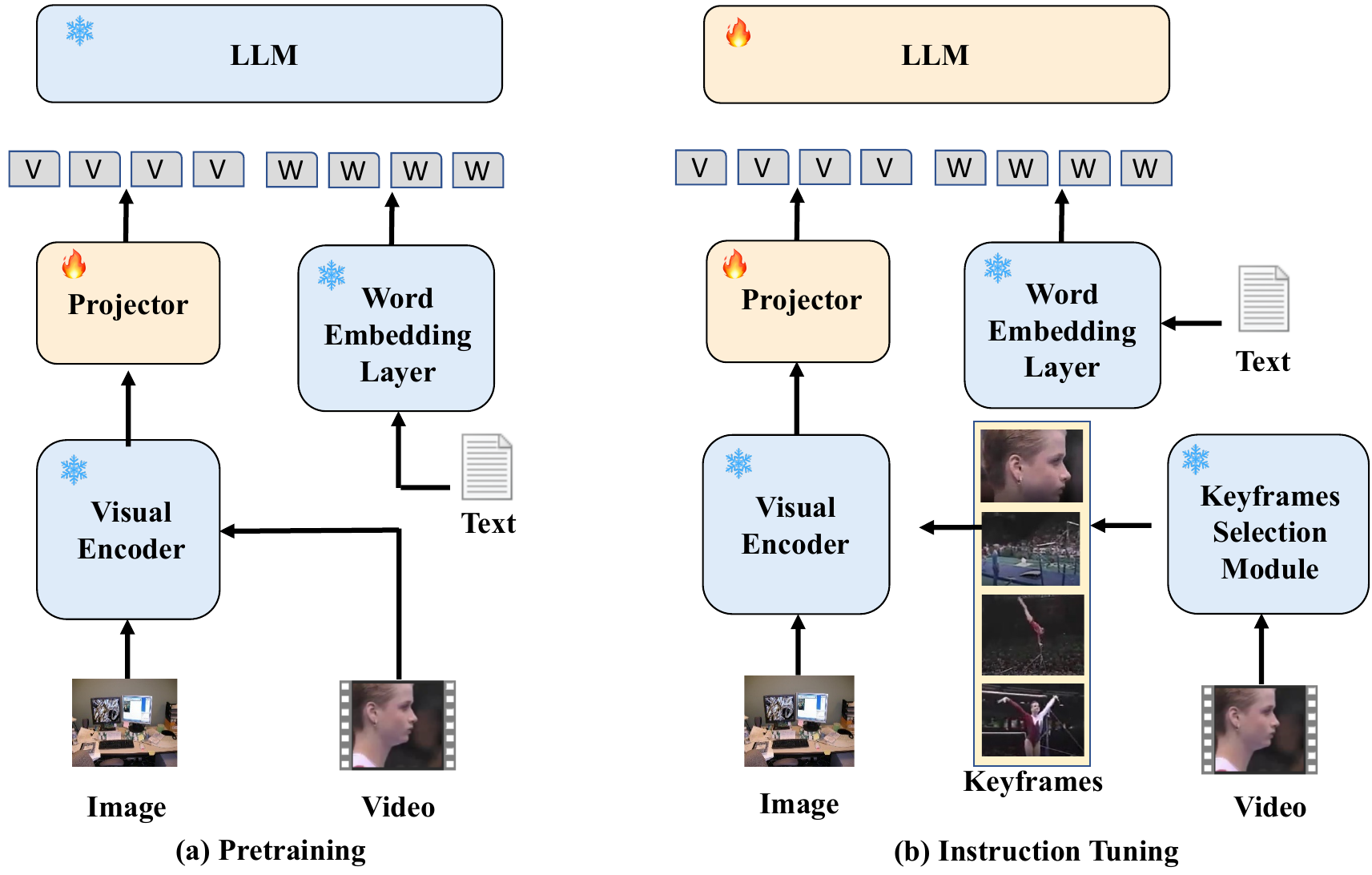}
  \caption{We use all the keyframe selection methods in the instruction tuning stage.}
  \label{fig:arcti}
  \vspace{1mm}
\end{figure*}
%


\subsubsection{Coarse Level Keyframe Selection}
In the coarse frame selection stage, to avoid selecting frames with too small frame spacing and to ensure sample diversity, a uniform sampling method is used to select a number of coarse frames (cn) frames. Specifically, cn is set to 32 frames in this work.
\subsubsection{Fine Level Keyframe Selection}
In the fine frame selection stage, we first consider the coarse-level selected video frames as a set \(\mathcal{F}\). We take the set \(\mathcal{F}\) as input and feed it to the image encoder, and each frame in \(\mathcal{F}\) gets a corresponding visual vector \(\mathbf{v}\). After that, we compute the similarity of these visual vectors \(\mathbf{v}\) with the word embedding to get the similarity score. The similarity score is calculated as follows: 
\[
score(\mathbf{v}_i, \mathbf{w})=\frac{\mathbf{v}_i \cdot \mathbf{w}}{\left\|\mathbf{v}_i\right\|\left\|\mathbf{w}\right\|}, \eqno{(1)}
\]
where $\mathbf{v}_i$ is the i-th visual embedding and $\mathbf{w}$ is the word embedding.



Next, we sort these similarity scores and select the top k with the highest scores. After that, we identify the corresponding video frames, select these video frames, and form a set of keyframes at the fine level. Finally, we recombine the collection of frames into a video in the original video's temporal order, so that for each dialog text, there is a unique counterpart consisting of keyframes.

The advantages of our approach are:
\begin{itemize}
    \item Compared to clustering methods, our method does not require additional parameter settings. Different settings of the hyperparameters have a relatively large impact on the effect of frame selection, which suggests that our approach is more robust.
    \item Compared to deep learning-based video summarization methods, our method does not require costly video pre-training, leading to higher efficiency.
    \item Compared to the other frame selection methods, our method selects keyframes that are more relevant to the content of the question-answer. Therefore, when the large model uses the keyframes and question-answer selected by our method in the training phase, it receives more accurate supervised information, which improves the understanding of the video. 
\end{itemize}

\subsection{CLIP-based Keyframe Selection for VideoLLM Training}
The CLIP-based Keyframe Selection method for training is illustrated in Fig. \ref{fig:arcti}. VideoLLMs leverages encoders like LanguageBind\cite{languagebind_zhu2023} and CLIP~\cite{clip} to select both spatial and temporal video features. This is accomplished by averaging frame-level features across the temporal and spatial dimensions. Then the features are projected and concatenate with word embeddings for LLMs to understand.
 
The entire training framework is divided into two stages: pre-training and instruction tuning. In the pre-training stage, we use video-text pairs to align vision and text. Similar to most other methods~\cite{video-llava, internvideo2}, we freeze the parameters of the large language model (LLM) and the visual encoder, training only the projector. This approach allows the projector to endow the LLM with video understanding capabilities without compromising its language abilities.

In the supervised instruction tuning stage, the model is tuned using video question-answering datasets. Previous VideoLLMs commonly use random or uniform frame selection~\cite{video-llava, internvideo, internvideo2}. In contrast, we introduce a keyframe extraction module. As mentioned in Section~\ref{sec:CLIP-based Keyframe Extraction}, we use CLIP-A and CLIP-QA to select more relevant frames. The pre-trained model is further fine-tuned using keyframes selected by CLIP-A and CLIP-QA to create high-quality text-video frame pairs. 

\subsection{CLIP-based Keyframe Selection for VideoLLM Inference}
During the testing (inference) phase, existing benchmarks typically provide questions about the video, which are then answered by VideoLLMs. Due to the presence of a large number of frames in a video, many are redundant or even interfere with video understanding. VideoLLMs struggle to process such massive amounts of frames effectively. To address this, previous VideoLLMs~\cite{video-llava, internvideo, internvideo2} commonly use uniform frame selection. However, this method does not focus on the frames relevant to the question. In contrast, we select keyframes based on the question. As mentioned in Section~\ref{sec:CLIP-based Keyframe Extraction}, we use CLIP-Q to leverage question information to select frames relevant to the question for inference. The selected frames are then used for video question-answering.

\section{Experiments}
In this section, we first introduce the experimental setups. We then aim to answer the following questions to verify the effectiveness, efficiency, and robustness of our proposed KeyframeLLM: \textbf{Q1}: Can our CLIP-A and CLIP-QA methods outperform uniform frame selection and other SoTA keyframe selection methods during the training stage? \textbf{Q2}: Can our CLIP-Q method further outperform uniform frame selection during the inference stage? \textbf{Q3}: How efficient and robust is our CLIP-based method compared to previous methods? \textbf{Q4}: Can our method generalize well across other model architectures? \textbf{Q5}: Can we visualize the advantages of our method?

\subsection{Experimental Settings}
\paragraph{\textbf{Datasets.}}
For training videos, we employ the same pre-training video datasets utilized by Video-LLaVA~\cite{video-llava}. Additionally, we incorporate the Video Instruction Dataset for Video Instruction Tuning from VideoChatGPT~\cite{video-chatgpt}, which provides a comprehensive resource of video question-answer pairs. This diverse dataset ensures robust training and instruction tuning of our models.

For inference and evaluation, we use well-established video datasets including ActivityNet, MSRVTT, MSVD, and TGIF. These datasets are consistent with those used for evaluation in VideoChatGPT~\cite{video-chatgpt}, providing a reliable basis for performance comparison and validation of our method.

\paragraph{\textbf{Models.}}
Our VideoLLM experiments utilize the SoTA framework, Video-LLaVA~\cite{video-llava}. For keyframe selection, we chose the pre-trained CLIP model with a patch size of 32 as the encoder due to its superior performance in aligning visual and textual data.

\paragraph{\textbf{Baselines.}}
We compare the performance of KeyVideoLLM with several baseline keyframe selection methods, including uniform frame selection, Katna, and DSNet. These baselines are selected due to their popularity and previous use in similar research, providing a robust comparative analysis for our proposed method.

\paragraph{\textbf{Settings.}}
For Video-LLaVA, we primarily use the hyperparameters from the official repository. For the CLIP model, we choose CLIP-ViT-B/32 to conduct keyframe selection. For the evaluation, we use LLaMA3 8B~\cite{llama} to rate our results. All experiments are conducted on an 8*A100 NVIDIA GPU machine with a 120-core CPU and 960GB of memory.

\subsection{Keyframe Selection for Training}\label{sec:Keyframe Extraction for Training}
To address \textbf{Q1}, we compare the performance of KeyVideoLLM (CLIP-A) and KeyVideoLLM (CLIP-QA) with other keyframe selection methods, including uniform selection (Baseline), Katna~\cite{katna}, and DSNet~\cite{DSNet}, during the supervised instruction tuning stage. These keyframe selection methods are used to select keyframes, which are then utilized to train the model. 
We employ the VideoLLM framework Video-LLaVA~\cite{video-llava} for our experiments in this section. For additional experiments, please refer to section \ref{app: More Experiments}.

The results of these comparisons are summarized in Table \ref{tab:train_clip}. As shown in Table \ref{tab:train_clip}, KeyVideoLLM (CLIP-QA) consistently outperforms other methods in terms of both score and accuracy across all datasets. This demonstrates that using frames related to both the question and answer can significantly enhance the performance of VideoLLMs during training.

KeyVideoLLM (CLIP-A) also achieved strong results, though slightly lower than KeyVideoLLM (CLIP-QA), indicating that the inclusion of answer information is beneficial. Additionally, incorporating more information (question) for keyframe selection yields better outcomes.

Katna and DSNet are keyframe selection methods that focus on general key information without specific relevance to the question. These methods do not consistently outperform the baseline, suggesting that non-question-aware keyframe selection methods do not provide a significant advantage over uniform selection.

\begin{table*}
\centering
\caption{Performance across different datasets in the training scenario. The values represent the score and accuracy (ACC). We use all keyframe selection methods only in the training stage and uniform frame selection in the inference stage.}
\begin{tabular}{lcccccccc}
\toprule
& \multicolumn{2}{c}{Activity} & \multicolumn{2}{c}{MSRVTT} & \multicolumn{2}{c}{MSVD} & \multicolumn{2}{c}{TGIF} \\
\cmidrule(r){2-3} \cmidrule(r){4-5} \cmidrule(r){6-7} \cmidrule(r){8-9}
Method & score & acc & score & acc & score & acc & score & acc \\
\midrule
Baseline & 2.18 & 0.45 & 2.38 & 0.55 & 2.90 & 0.66 & 2.32 & 0.52 \\
Katna & 2.18 & 0.45 & 2.39 & 0.55 & 2.93 & 0.67 & 2.30 & 0.51 \\
DSNET & 2.17 & 0.45 & 2.38 & 0.55 & 2.87 & 0.65 & 2.32 & 0.51 \\
CLIP-A & 2.20 & 0.46 & 2.39 & 0.55 & \textbf{2.94} & \textbf{0.67} & \textbf{2.36} & \textbf{0.52} \\
CLIP-QA & \textbf{2.21} & \textbf{0.46} & \textbf{2.40} & \textbf{0.56} & \textbf{2.94} & \textbf{0.67} & \textbf{2.36} & \textbf{0.52} \\
\bottomrule
\end{tabular}
\label{tab:train_clip}
\end{table*}

\subsection{Keyframe Selection for Inference}\label{sec:Keyframe Extraction for Inference}

\begin{table*}
\centering
\caption{Performance across different datasets in the inference scenario. The values represent the score and accuracy (ACC). We use the same keyframe selection methods as in Table \ref{tab:train_clip} and CLIP-Q for inference.}
\begin{tabular}{lcccccccc}
\toprule
& \multicolumn{2}{c}{Activity} & \multicolumn{2}{c}{MSRVTT} & \multicolumn{2}{c}{MSVD} & \multicolumn{2}{c}{TGIF} \\
\cmidrule(r){2-3} \cmidrule(r){4-5} \cmidrule(r){6-7} \cmidrule(r){8-9}
Method & score & acc & score & acc & score & acc & score & acc \\
\midrule
Baseline & 2.21 & 0.46 & 2.38 & 0.55 & 2.92 & 0.66 & 2.33 & 0.51 \\
Katna & 2.22 & 0.46 & 2.39 & 0.55 & 2.97 & 0.68 & 2.30 & 0.51 \\
DSNET & 2.17 & 0.45 & 2.38 & 0.55 & 2.95 & 0.67 & 2.32 & 0.51 \\
CLIP-A & \textbf{2.22} & \textbf{0.46} & 2.39 & 0.55 & \textbf{2.98} & \textbf{0.68} & \textbf{2.36} & \textbf{0.52} \\
CLIP-QA & \textbf{2.22} & \textbf{0.46} & \textbf{2.40} & \textbf{0.56} & \textbf{2.98} & \textbf{0.68} & \textbf{2.36} & \textbf{0.52} \\
\bottomrule
\end{tabular}
\label{tab:inf_clip}
\end{table*}
To address \textbf{Q2}, in this section, we compare the performance of KeyVideoLLM (CLIP-Q) with uniform selection in the inference scenario. Using the models trained in Section \ref{sec:Keyframe Extraction for Training}, we fix the model and apply KeyVideoLLM (CLIP-Q) to select frames relevant to the question for VideoLLMs inference. The results of KeyVideoLLM (CLIP-Q) are summarized in Table \ref{tab:inf_clip}. We then compare our approach with uniform frame selection for inference, as shown in Table \ref{tab:train_clip}.

As illustrated in Table \ref{tab:inf_clip}, KeyVideoLLM (CLIP-QA) continues to outperform KeyVideoLLM (CLIP-A), Katna, DSNet, and the baseline. This consistency indicates the effectiveness of our method.
Additionally, KeyVideoLLM demonstrates superior performance compared to uniform keyframe selection during inference. By simply changing the keyframes used for inference, our model's performance improves compared to the results shown in Table \ref{tab:train_clip}.

Notably, the performance increase is substantial for the ActivityNet and MSVD datasets, which consist of longer videos. Longer videos present a greater challenge for uniform selection to capture frames relevant to the question, hence the more significant performance boost with our method. Conversely, the improvement is relatively lower for the TGIF and MSRVTT datasets, which contain shorter videos.

Furthermore, employing keyframe selection during both the training and inference stages enables the achievement of SoTA results. By focusing on the most relevant frames, KeyVideoLLM reduces data redundancy and enhances effectiveness, leading to superior model performance.

\subsection{Efficiency and Robustness Analysis}
To address \textbf{Q3}, we analyze the efficiency and robustness of KeyVideoLLM by comparing it with other keyframe selection methods and a baseline method (without keyframe selection). Our analysis focuses on three key aspects: compression ratio, selection success rate, and selection speed.

\paragraph{1. Highest Compression Ratio}
To quantify the compression ratio achieved by KeyVideoLLM, we use the following formula:

\[
\text{Compression Ratio} = \frac{S_{\text{orig}}}{S_{\text{comp}}},
\eqno{(2)}
\]

where \(S_{\text{orig}}\) represents the total size of the video data before applying keyframe selection, and \(S_{\text{comp}}\) represents the total size of the video data after applying keyframe selection.

\begin{figure}[ht]
  \includegraphics[width=8cm]{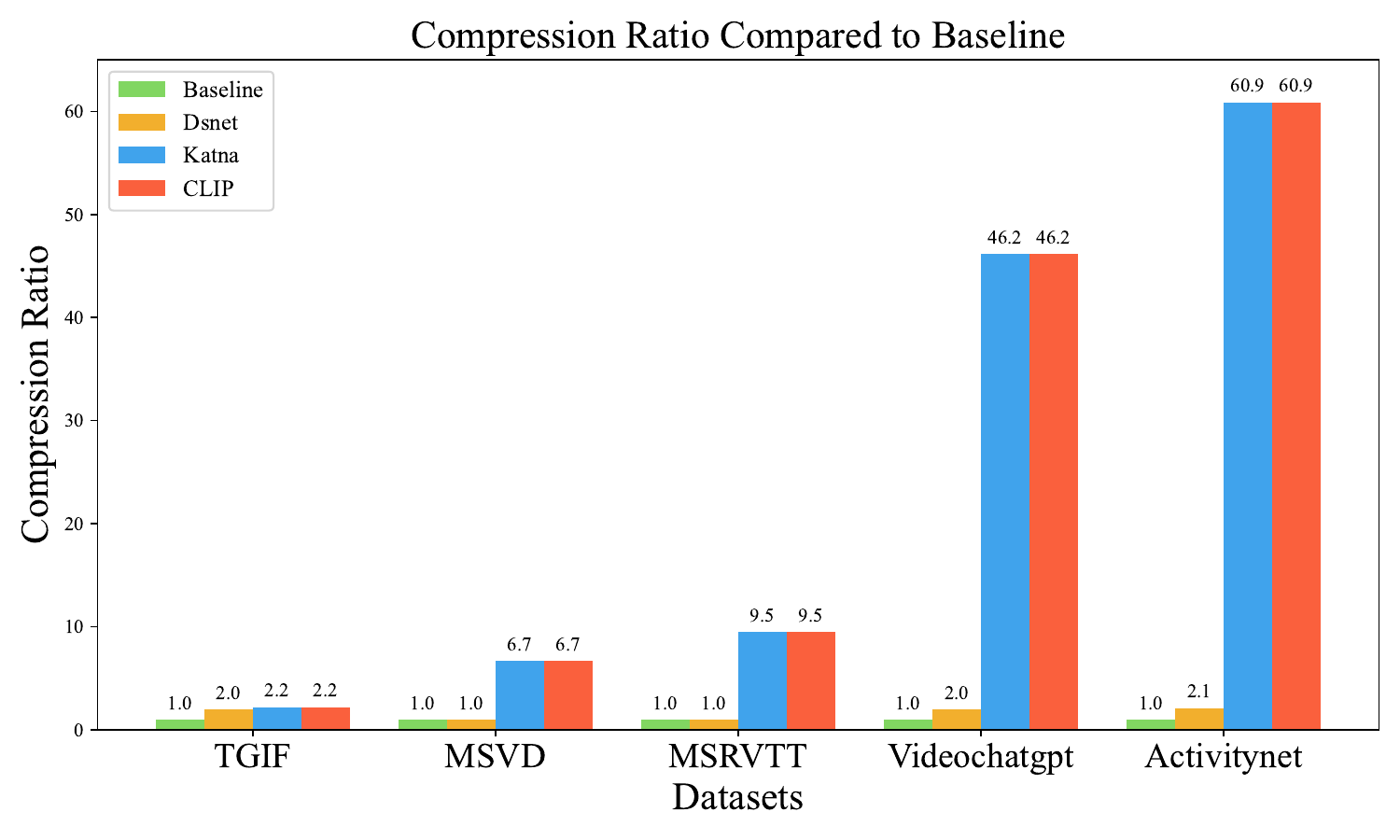}
  \caption{Compression Ratios of Various Methods on Different Datasets. The graph illustrates the compression ratios achieved by our model KeyVideoLLM compared to Katna and DSNet across five different datasets. Higher ratios indicate more efficient compression, demonstrating the superior computational and storage efficiency of our approach.}
  \label{fig:Compress_Ratio}
\end{figure}

A higher compression ratio indicates a more efficient compression method, as it means the model can reduce the data size more significantly while maintaining the necessary information for effective video question-answering. As shown in Figure \ref{fig:Compress_Ratio}, KeyVideoLLM achieves the highest compression ratios compared to Katna and DSNet across five different datasets. The graph illustrates that our model (CLIP-A and CLIP-QA) significantly reduces data size (up to 60 times) while preserving essential information, demonstrating superior computational and storage efficiency. Higher compression ratios indicate more efficient data usage, making our approach highly effective for large-scale video processing tasks.

\paragraph{2. Highest Success Rate}
Our method also achieves the highest selection success rate across all datasets, as shown in Table \ref{tab:Success_Rate}. It consistently outperforms Katna and DSNet, demonstrating its reliability and accuracy in various video scenarios.
\begin{table}
  \centering
    \caption{
    Success rate of different keyframe selection methods across various datasets. Selecting fewer than 8 frames is considered a failure.
  }
  \begin{tabular}{lccccc}
    \hline
     & \textbf{Datasets}  & \textbf{Katna} & \textbf{DSNet}  & \textbf{Ours} & \\
    \hline
    \multirow{5}{*}{\textbf{Success Rate}} & VideoChatGPT & 99.8\% & 57.8\% & \textbf{100.0\%} \\
    & ActivityNet & 99.1\% & 61.1\% & \textbf{100.0\%} \\
    & MSRVTT & <5\% & <5\% & \textbf{100.0\%} \\
    & MSVD & <5\% & <5\% & \textbf{100.0\%} \\
    & TGIF & <5\% & <5\% & \textbf{100.0\%} \\
    \hline
  \end{tabular}
  \label{tab:Success_Rate}
\end{table}

\paragraph{3. Fastest Selection Speed}
Finally, our method boasts the fastest keyframe selection speed, as detailed in Table \ref{tab:Extract_Speed}. The selection speed (measured in seconds per video) highlights the efficiency of KeyVideoLLM in processing large volumes of video data quickly. This speed advantage further solidifies the practicality of our approach in real-world applications where time efficiency is critical.
\begin{table}
  \centering
    \caption{
    Average selection speed (in seconds per video) of different keyframe selection methods across various datasets. For MSRVTT, MSVD, and TGIF, Katna and DSNet failed (<5\% Success Rate) to select 8 frames for short videos, so their selection times are not included for these datasets.
  }
  \begin{tabular}{lcccc}
    \hline
    & \textbf{Datasets}  & \textbf{Katna} & \textbf{DSNet}  & \textbf{Ours} \\
    \hline
    \multirow{5}{*}{\textbf{Selection Speed}} & VideoChatGPT & 50 & 3.5 & \textbf{0.25}\\
    & ActivityNet & 50 & 3.2 & \textbf{0.25}\\
    & MSRVTT & - & - & \textbf{0.24}\\
    & MSVD & - & - & \textbf{0.24}\\
    & TGIF & - & - & \textbf{0.24}\\
    \hline
  \end{tabular}
  \label{tab:Extract_Speed}
\end{table}

\subsection{Generalizability of KeyframeLLM}\label{app: More Experiments}
To address \textbf{Q4}, following the experiment results in \ref{sec:Keyframe Extraction for Training}. In this section, we provide additional experimental results to further validate the effectiveness of KeyframeLLM. Specifically, we investigate the impact of using different encoder architectures for our VideoLLM. In the main experiments, we utilized the VideoLLM Video-LLaVA~\cite{video-llava} as our model. To explore the robustness and generalizability of our keyframe selection methods, we conducted supplementary experiments by replacing the encoder architecture with CLIP~\cite{clip}. 

We followed the same experimental setup as described in Section~\ref{sec:Keyframe Extraction for Training} and Section~\ref{sec:Keyframe Extraction for Inference}. We used the CLIP encoder to process the keyframes selected by our methods and trained the VideoLLM accordingly. The datasets used for training and evaluation remain unchanged: VideoChatGPT is used for training, while ActivityNet, MSRVTT, MSVD, and TGIF are used for evaluation.

\begin{table}
\centering
\small
\caption{Performance across different datasets in the training scenario. The values represent the score and accuracy (acc).}
\begin{tabular}{lcccccccc}
\toprule
& \multicolumn{2}{c}{Activity} & \multicolumn{2}{c}{MSRVTT} & \multicolumn{2}{c}{MSVD} & \multicolumn{2}{c}{TGIF} \\
\cmidrule(r){2-3} \cmidrule(r){4-5} \cmidrule(r){6-7} \cmidrule(r){8-9}
Method & score & acc & score & acc & score & acc & score & acc \\
\midrule
Baseline & 2.05 & 0.43 & 2.33 & 0.54 & 2.85 & 0.65 & 2.15 & 0.47 \\
Katna & 2.13 & 0.44 & 2.33 & 0.54 & 2.84 & 0.65 & 2.18 & 0.48 \\
DSNET & 2.10 & 0.44 & 2.28 & 0.53 & 2.77 & 0.63 & 2.17 & 0.48 \\
CLIP-A & 2.15 & 0.44 & 2.37 & \textbf{0.55} & 2.89 & \textbf{0.66} & 2.19 & 0.48 \\
CLIP-QA & \textbf{2.16} & \textbf{0.44} & \textbf{2.38} & \textbf{0.55} & \textbf{2.90} & \textbf{0.66} & \textbf{2.24} & \textbf{0.49} \\
\bottomrule
\end{tabular}
\label{tab:train_clip_clip_enc}
\end{table}

\begin{table}
\centering
\small
\caption{Performance across different datasets in the training scenario. The values represent the score and accuracy (acc).}
\begin{tabular}{lcccccccc}
\toprule
& \multicolumn{2}{c}{Activity} & \multicolumn{2}{c}{MSRVTT} & \multicolumn{2}{c}{MSVD} & \multicolumn{2}{c}{TGIF} \\
\cmidrule(r){2-3} \cmidrule(r){4-5} \cmidrule(r){6-7} \cmidrule(r){8-9}
Method & score & acc & score & acc & score & acc & score & acc \\
\midrule
Baseline & 2.06 & 0.43 & 2.33 & 0.54 & 2.86 & 0.65 & 2.15 & 0.47 \\
Katna & 2.16 & 0.45 & 2.33 & 0.54 & 2.84 & 0.65 & 2.22 & 0.49 \\
DSNET & 2.12 & 0.44 & 2.28 & 0.53 & 2.79 & 0.64 & 2.18 & 0.48 \\
CLIP-A & 2.16 & \textbf{0.45} & 2.37 & \textbf{0.55} & \textbf{2.94} & \textbf{0.68} & 2.19 & 0.48 \\
CLIP-QA & \textbf{2.17} & \textbf{0.45} & \textbf{2.38} & \textbf{0.55} & \textbf{2.94} & \textbf{0.68} & \textbf{2.24} & \textbf{0.49} \\
\bottomrule
\end{tabular}
\label{tab:inf_clip_clip_enc}
\end{table}

\subsection{Qualitative Evaluation}
\begin{figure}[ht]
  \includegraphics[width=8cm]{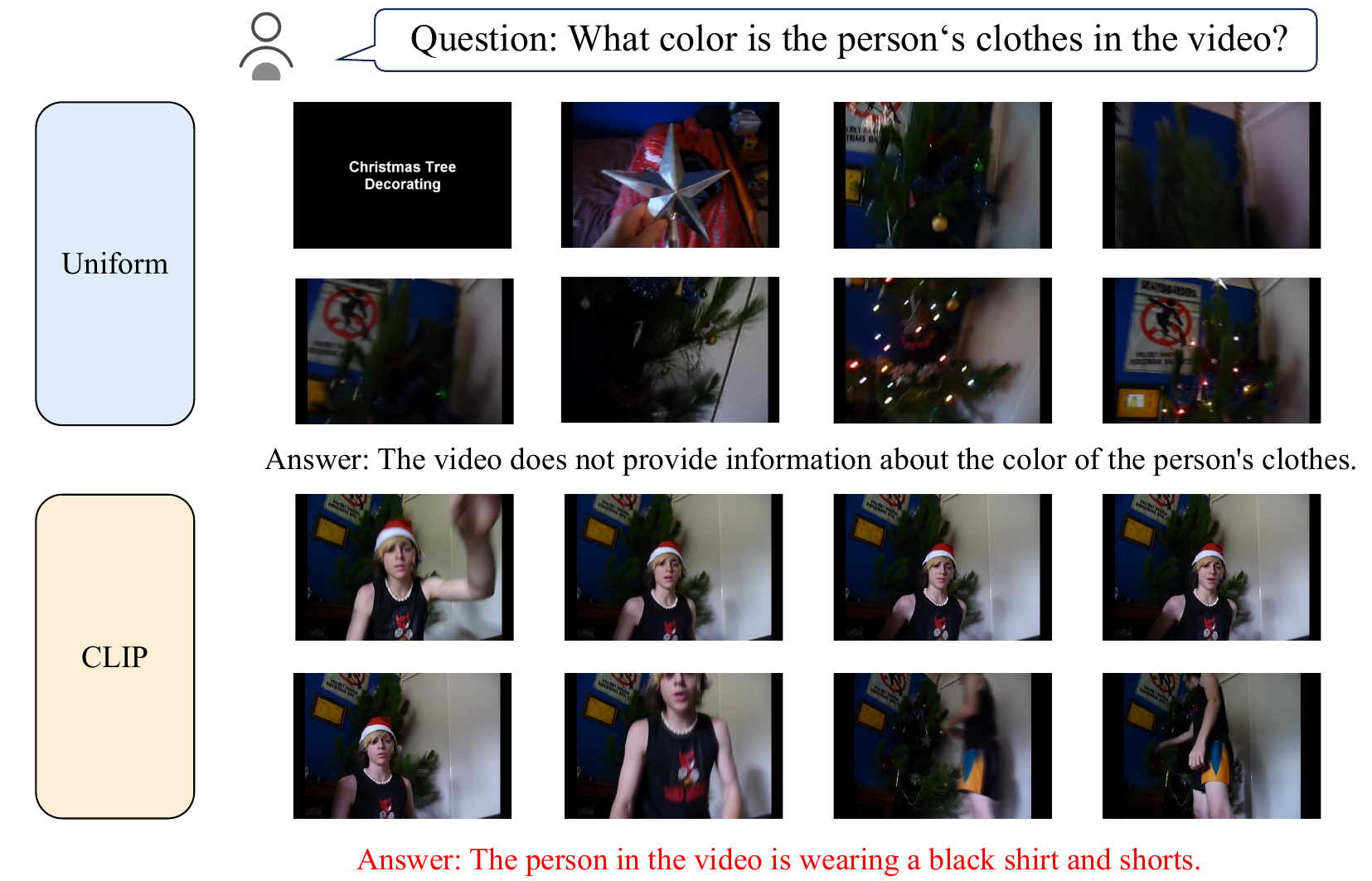}
  \caption{Comparison between uniform frame selection and CLIP-based frame selection for video question-answering. The uniform selection often fails to capture relevant information, while the CLIP-based selection consistently identifies frames that are pertinent to the question about the color of the person's clothes.}
  \label{fig:CLIP_Inference}
\end{figure}

To address \textbf{Q5}, in this section, we provide a qualitative evaluation of our method to demonstrate its effectiveness in video selection tasks. As shown in Figure \ref{fig:Face}, we present a comparison between the baseline response, which is generated by uniform frame selection, and the response generated by our KeyVideoLLM (CLIP-Q) model.

In Figure \ref{fig:Face}, the question is: "Is the person in the white coat wearing a hat?" The baseline model, due to uniform frame selection, captures only a vague frame, leading to the incorrect response: "Yes, the person in the white coat is wearing a hat." In contrast, our KeyVideoLLM (CLIP-Q) selects a clear and relevant frame, allowing the model to correctly identify that the person in the white coat is not wearing a hat, thus providing the accurate response: "No, the person in the white coat is not wearing a hat."

We provide an additional qualitative evaluation to further demonstrate our model's effectiveness in video selection tasks. As shown in Figure \ref{fig:CLIP_Inference}, we compare the baseline response, generated by uniform frame selection, with the response generated by our KeyVideoLLM (CLIP-Q) model.

In the example depicted, the question is: "What color are the person's clothes in the video?" The baseline model, which selects frames uniformly, fails to provide relevant information, resulting in the response: "The video does not provide information about the color of the person's clothes." In contrast, our KeyVideoLLM (CLIP-Q) model accurately identifies keyframes relevant to the question, providing the correct response: "The person in the video is wearing a black shirt and shorts."

These qualitative analyses underscore the superior performance of KeyVideoLLM in understanding and selecting relevant keyframes for accurate video question-answering. The model's ability to leverage Answer and Question-Answer pairs for keyframe selection significantly enhances its accuracy and reliability compared to traditional methods.

\section{Conclusion}
VideoLLMs are emerging as powerful deep learning models designed for video question-answering tasks. Efficient, robust, and effective keyframe selection algorithms are essential for training VideoLLMs, but they remain challenging due to their inherent complexity. 
This paper presents KeyVideoLLM, a new approach to select keyframes for VideoLLMs by leveraging the text-video frames similarity scores. 
Experimental results on diverse datasets indicate that KeyVideoLLM significantly improves the performance of VideoLLMs during both the training and inference stages. Furthermore, it consistently outperforms the compared baseline methods in terms of efficiency, effectiveness, and robustness.


\appendix

\newpage
\bibliographystyle{ACM-Reference-Format}
\bibliography{main}

\end{document}